\documentclass[twocolumn]{article}
\usepackage[a4paper, margin=1.5cm]{geometry}
\usepackage{cite}
\usepackage{amsmath,amssymb,amsfonts}
\usepackage{graphicx}
\usepackage{textcomp}
\usepackage{bm}

\usepackage{verbatim}
\usepackage{algorithm}
\usepackage{algpseudocode}
\usepackage{multirow}
\usepackage{flushend}
\usepackage{multicol}

\usepackage{url}
\usepackage{caption}
\usepackage{newfloat}
\DeclareFloatingEnvironment[name=Listing]{listing}

\title{Error-Driven Prompt Optimization for Arithmetic Reasoning
\\\normalsize \textit{A Code Generation Approach Using On-Premises Small Language Models on Tabular Data}
}

\author{
Árpád Pándy, Róbert Lakatos, András Hajdu\\
\small
Deptartment of Data Science \& Visualization, Faculty of Informatics, University of Debrecen
}

\begin{document}



\maketitle

\begin{abstract}
Recent advancements in artificial intelligence have sparked interest in industrial agents capable of supporting analysts in regulated sectors, such as finance and healthcare, within tabular data workflows. A key capability for such systems is performing accurate arithmetic operations on structured data while ensuring sensitive information never leaves secure, on-premises environments. Here, we introduce an error-driven optimization framework for arithmetic reasoning that enhances a Code Generation Agent (CGA), specifically applied to on-premises small language models (SLMs). Through a systematic evaluation of a leading SLM (Qwen3 4B), we find that while the base model exhibits fundamental limitations in arithmetic tasks, our proposed error-driven method, which clusters erroneous predictions to refine prompt-rules iteratively, dramatically improves performance, elevating the model's accuracy to 70.8\%. Our results suggest that developing reliable, interpretable, and industrially deployable AI assistants can be achieved not only through costly fine-tuning but also via systematic, error-driven prompt optimization, enabling small models to surpass larger language models (GPT-3.5 Turbo) in a privacy-compliant manner.

\medskip
\noindent\textbf{Keywords:}  Arithmetic calculation, code-generation agent, generative large-language models, prompt engineering, tabular data.
\end{abstract}



\section{Introduction}
\label{sec:intro}

Tabular data remains a cornerstone of knowledge representation in domains such as finance, healthcare, and the natural sciences. Yet their automated interpretation presents persistent challenges for language models \cite{lu2025large} \cite{dong2024large}. Early work has revealed that large language models (LLMs) perform impressively in natural language understanding. Still, their arithmetic reasoning is fragile: they often produce inconsistent or even wrong numerical outputs when faced with structured data. To overcome this limitation, our previous study \textit{(Arithmetic-aware question-answering on tabular data using a large language model-based code generation agent)} \cite{pandy_ines_2025} reframed question answering as a code-generation task, in which the model produces executable programs that perform the required data selection and arithmetic operations. When coupled with table restructuring and domain-specific rule injection, this hybrid framework—the Code Generation Agent (CGA)—achieved dramatic gains, raising exact-match accuracy on financial benchmarks from below 30\% to nearly 80\%. 

Despite these advances, the reliance on API-based large models posed a critical barrier for regulated sectors, where sensitive data cannot leave secure environments. A subsequent study \cite{closingpandy} investigated small language models (SLMs) in the 1–7B parameter range, which can run entirely on-premises while requiring more modest computational resources. Strikingly, the combination of CGA, table restructuring, and prompt simplification enabled certain SLMs—notably Qwen3 4B \cite{qwen3technicalreport} —to exceed GPT-3.5 Turbo in numerical accuracy, delivering both interpretability and privacy preservation. These results pointed to the feasibility of local, resource-efficient agents capable of robust tabular reasoning.

However, key questions remained: how to extend such agents beyond hand-crafted rules, which often fail across model scales, and how to ground prompting strategies in a theoretical framework that balances rule number and type. 

We present an error-driven framework that clusters mistaken predictions to derive domain-specific rules, iteratively refining reasoning prompts so the agent “learns from its mistakes” without costly fine-tuning. We formalize an optimal rule set ($K_{\text{opt}}$) balancing informativeness and cognitive load, explaining the performance curve. This method boosts Qwen3 4B to 70.82\% accuracy, surpassing GPT-3.5 Turbo while preserving full data sovereignty.

In this Article, we integrate CGA's methodological innovations, small language models' efficiency, and error-driven rule induction's systematic power into a cohesive blueprint for interpretable, auditable, and industrially deployable agents. We outline in Section \ref{sec:dataset} the dataset and the prompt extending algorithm \ref{sec:methodology}. We describe the implementation details in Section \ref{details} and Experiment \ref{sec:experiments}. Present results and theoretical insights \ref{sec:results}, concluding \ref{sec:conclusion} with a best-practice foundation for trustworthy tabular reasoning AI.

\section{Dataset}
\label{sec:dataset}


Although the number of available datasets containing QA test cases on tabular data with arithmetic calculations is limited, three datasets can be considered for this task: WikiTableQuestions, OTT-QA, and TAT-QA. 
We chose the TAT-QA dataset because it provides the calculation derivation for each question as metadata. It categorizes questions by reasoning type, an essential feature for testing a solution designed for arithmetic calculations. 
Dataset WikiTableQuestions \cite{pasupat2015wikitable} contains derivation annotations only for a limited set of questions, and no answer categorization is provided. Dataset OTT-QA \cite{chen2021ottqa} lacks both of these features. 

Our research used the "A Question Answering Benchmark on a Hybrid of Tabular and Textual Content in Finance" (TAT-QA) dataset \cite{zhu_tat-qa_2021} to evaluate the performance of the models. This data set contains financial reports in tabular format. It was designed to evaluate complex, hybrid QA tasks. Each financial report has multiple related text paragraphs and questions. The questions can be grouped by answer type. The answer can be a span or a multipart span of text from the table or the related paragraphs, a number resulting from an arithmetic calculation, or a counter, if the question is about counting data items.

\begin{figure}[ht!]
    \centering
    \includegraphics[width=1.0\linewidth]{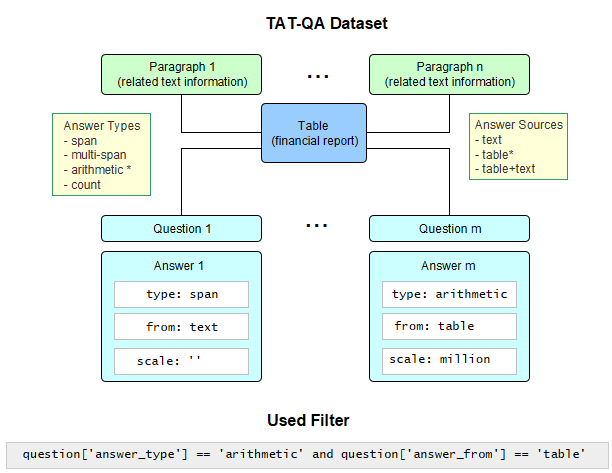}
    \caption{Structure of the TAT-QA dataset and the applied filter.}
    \label{fig:dataset}
\end{figure}

As our research focused on calculations on tabular data, we applied a filtering pipeline in the same way described in \cite{pandy_ines_2025}. The filter selected questions where the answer type was arithmetic calculations, which can be performed on the tabular data, without the need for the text in related paragraphs. Using these filters, the number of tables was reduced from 278 to 215, and the number of questions was reduced from 1668 to 497. The structure of the dataset and the applied filter are shown in Figure \ref{fig:dataset}.

The reference values are composed of two different parts: the predicted value and its scale. The scale can be one billion, million, thousand, or percent. It is reasonable to distinguish these two parts, as they can usually be found in different locations within the table or inferred from the question. An example of the former is when the financial report contains millions of dollars, as indicated by the table headers. An example of the latter is when the question is about a percentage, because in this case the result scale must be "percent" by definition.

\section{Methodology}
\label{sec:methodology}

\begin{figure}[!ht]
    \centering    
    \includegraphics[width=0.9\linewidth]{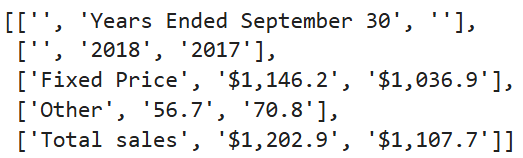}
    \caption{Original table format.}
    \label{fig:origtable}
\end{figure}

\begin{figure}[!ht]
    \centering
    \includegraphics[width=0.9\linewidth]{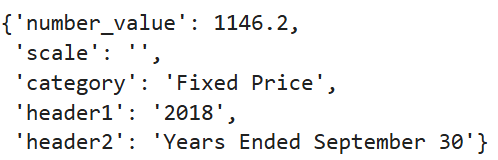}
    \caption{A data item after restructuring.}
    \label{fig:value_item}
\end{figure}

In this section, we show a method for extending the capabilities of the Code Generation Agent \cite{pandy_ines_2025}. 
In the cited research, we developed three improvement methods on arithmetic reasoning on tabular data.  The first is restructuring the table to a set of annotated values. The original table format can be seen in Figure \ref{fig:origtable}, while the restructured table is shown in Figure \ref{fig:value_item}. 

\begin{figure}[!ht]
    \centering
    \includegraphics[width=0.8\linewidth]{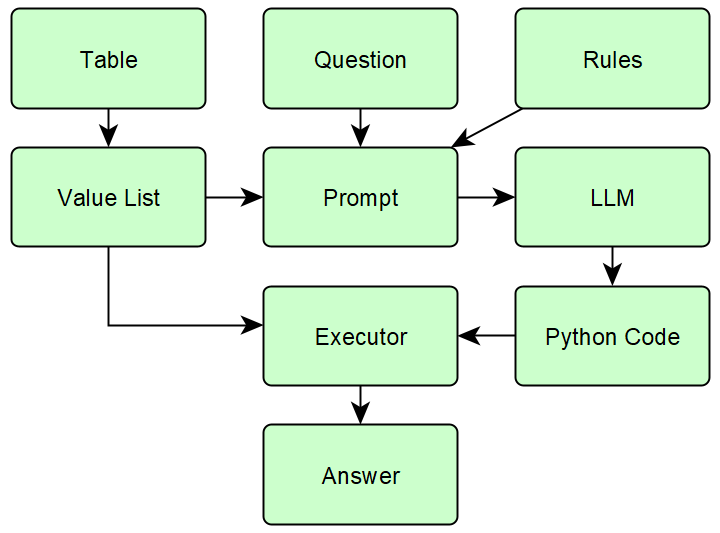}
    \caption{Overview of the process in Code Generation Agent.}
    \label{fig:flow}
\end{figure}

The second improvement method was the introduction of Code Generation Agent. This LLM based agent generates a Python code based on the table and the question, and executes the generated code. The third improvement method was the application of domain specific prompt rules. These rules are specific for the target domain / dataset. In this paper we show a method how these rules can be formulaated in a semi-automated way. The whole process is shown in Figure \ref{fig:flow}.

In our previous work \cite{closingpandy}, we examined the possibilities of applying CGA to SLMs, created a benchmark on selected small models, using the TAT-QA dataset, and demonstrated the effectiveness of denser prompts on small language models. 
We summarize the results of that research in the tables below. 
We used Exact Match (EM) as a metric to compare predicted and ground truth answer. The EM score for a test instance $(q,c,y)$ with model prediction $\hat{y}$ is defined as
\[
\text{EM}(q,c,y) = 
\begin{cases}
1 & \text{if } \hat{y} = y, \\
0 & \text{otherwise.}
\end{cases}
\]

The effects of different improvement techniques on small and large models are shown in Table \ref{tab:featurecomp}. Table \ref{tab:benchmark} shows the performance of selected small models. As a summary, we can state that the performance of the smallest models is close to zero, even when applying improvement methods; in the 3-4B size range, the performance can be improved significantly by allying CGA, compared to the reference large language model; and the performance of the top performer in this group achieves the performance of the reference model. Table \ref{tab:simplprompts} shows that with a simpler prompt, almost 60\% exact match can be achieved. 

\begin{table}[ht!]
\caption{Feature Performance comparison between model Gemma 3n E4B and GPT-3.5-Turbo}
\label{tab:featurecomp}
\centering
\small
\begin{tabular}{|l|l|l|}
\hline
Model              & \multicolumn{1}{c|}{Gemma 3n E4B}                                                                                     & \multicolumn{1}{c|}{GPT-3.5-Turbo}                                                                                    \\ \hline
                   & \multicolumn{1}{c|}{EM}               &  \multicolumn{1}{c|}{EM}                \\ \hline
Naive              & \multicolumn{1}{c|}{5.03\%}           &  \multicolumn{1}{c|}{29.38\%}           \\ \hline
Restructuring      & \multicolumn{1}{c|}{2.62\%}           &  \multicolumn{1}{c|}{21.33\%}           \\ \hline
Codegen Agent      & \multicolumn{1}{c|}{7.44\%}           &  \multicolumn{1}{c|}{36.42\%}           \\ \hline
CGA + Res.         & \multicolumn{1}{c|}{\textbf{29.78\%}} &  \multicolumn{1}{c|}{52.62\%}           \\ \hline
CGA + Res. + Rules & \multicolumn{1}{c|}{21.33\%}          &  \multicolumn{1}{c|}{\textbf{59.99\%}}  \\ \hline
\end{tabular}
\end{table}

\begin{table}[ht!]
\caption{Benchmark of model comparison with two setups: the naive model and when CGA and table restructuring are applied in the pipeline. 
}
\label{tab:benchmark}
\centering
\small
\begin{tabular}{|l|l|l|l|}
\hline
\multicolumn{1}{|c|}{\%}    & \multicolumn{1}{c|}{naive}                                                                                      & \multicolumn{1}{c|}{CGA + Res.}                                                                                 & \multicolumn{1}{c|}{Change}                                                                                     \\ \hline
\multicolumn{1}{|c|}{Model} & \multicolumn{1}{c|}{EM}             &  \multicolumn{1}{c|}{EM}             &  \multicolumn{1}{c|}{EM}              \\ \hline
Gemma 3n E4B                 & \multicolumn{1}{c|}{5.03}           & \multicolumn{1}{c|}{29.78}          & \multicolumn{1}{c|}{24.75}          
\\ \hline
Gemma 3 1B                   & \multicolumn{1}{c|}{0.00}           &  \multicolumn{1}{c|}{1.66}           & \multicolumn{1}{c|}{1.66}           
\\ \hline
Gemma 3 4B                   & \multicolumn{1}{c|}{3.02}           &  \multicolumn{1}{c|}{34.81}          &  \multicolumn{1}{c|}{31.79}          
\\ \hline
Llama 3.2 1B                & \multicolumn{1}{c|}{0.00}           &  \multicolumn{1}{c|}{1.01}           &   \multicolumn{1}{c|}{1.01}           
\\ \hline
Llama 3.2 3B                & \multicolumn{1}{c|}{0.00}           &  \multicolumn{1}{c|}{30.58}          &  \multicolumn{1}{c|}{30.58}          
\\ \hline
Mistral 7B                  & \multicolumn{1}{c|}{0.60}           & \multicolumn{1}{c|}{17.91}          &  \multicolumn{1}{c|}{17.31}          
\\ \hline
Qwen3 4B                  & \multicolumn{1}{c|}{4.83}           &  \multicolumn{1}{c|}{\textbf{53.72}}          &  \multicolumn{1}{c|}{\textbf{48.89}} \\ \hline
GPT-3.5 Turbo                   & \multicolumn{1}{c|}{\textbf{29.38}} &  \multicolumn{1}{c|}{52.62} & \multicolumn{1}{c|}{23.24} 
\\ \hline
\end{tabular}
\end{table}

\begin{table}[ht!]
\caption{Comparison of the models' performance using original and simplified prompts in the Code Generation Agent. 
}
\label{tab:simplprompts}
\centering
\small
\begin{tabular}{|l|ll|ll|ll|}
\hline
\multicolumn{1}{|c|}{\%}    & \multicolumn{1}{c|}{\begin{tabular}[c]{@{}c@{}}Complex\\ Prompt\end{tabular}}                                       & \multicolumn{1}{c|}{\begin{tabular}[c]{@{}c@{}}Simple\\ Prompt\end{tabular}}                                        & \multicolumn{1}{c|}{Change}                                                                                         \\ \hline
\multicolumn{1}{|c|}{Model} & \multicolumn{1}{c|}{EM}             &  \multicolumn{1}{c|}{EM}             & \multicolumn{1}{c|}{EM}              \\ \hline
Gemma 3n E4B                & \multicolumn{1}{c|}{29.78}                                                           & \multicolumn{1}{c|}{42.05}                                                                & \multicolumn{1}{c|}{12.27} 
                                                                          \\ \hline
Gemma3 4B                   & \multicolumn{1}{c|}{34.81}          & \multicolumn{1}{l|}{42.66}          &  \multicolumn{1}{c|}{7.85}           
\\ \hline
Llama 3.2 3B                & \multicolumn{1}{c|}{30.58}          & \multicolumn{1}{c|}{22.33}          &  \multicolumn{1}{c|}{-8.25}           
\\ \hline
Mistral 7B                  & \multicolumn{1}{c|}{17.91}          &  \multicolumn{1}{c|}{41.45}          &  \multicolumn{1}{c|}{\textbf{23.54}}           
\\ \hline
Qwen3 4B                    & \multicolumn{1}{c|}{\textbf{53.72} }&  \multicolumn{1}{c|}{\textbf{59.96}} &  \multicolumn{1}{c|}{6.24} 
\\ \hline
\end{tabular}
\end{table}

In the following, 
we formulate our hypotheses on introducing prompt rules into the prompt \ref{sec:promptdirec}, and finally, we describe the Prompt Extending Algorithm in pseudo-code in Algorithm \ref{alg:pea}.

\textit{Domain-specific Prompt Rules:}
Domain-specific prompt rules/instructional constraints (DSPR) are a set of instructions added to the prompt to enhance the model's performance. We will describe them in detail in Section \ref{sec:experiments}.

\subsection{Task Decomposition Model}
\label{sec:taskdm}

In this section, we give a formal description of the task decomposition method. The method is described in detail in \cite{pandy_ines_2025} and \cite{closingpandy}. Here, we aim to formalize this method and provide a theoretical framework for further improvements. 

The experimental results show that in the \emph{naive} approach
\[
f_{\text{naive}} : (Q, C) \to Y
\]
the probability of producing a correct answer
\[
P(A_{\text{correct}} \mid Q, C)
\]
It is low, since the model must perform semantic interpretation and arithmetic computation in a single step.

The \emph{CGA} (decomposition-based) approach divides the task into two steps:
\[
f_{\text{CGA}} : (Q, C') \to P, \quad \text{e}: P \to Y
\]
where $C'$ is the restructured table, $P$ is the generated code, and \texttt{e} is a deterministic executor.

In this case, the overall success probability can be approximated as:
\[
P(A_{\text{correct}} \mid Q, C') \approx P(P_{\text{correct}} \mid Q, C'),
\]
since $\texttt{e}$ is error-free.  
Measurements indicate that
\[
P_{\text{CGA}} - P_{\text{naive}} \gg 0,
\]
Decomposition and code generation significantly improve accuracy, especially when combined with table restructuring and the addition of prompt rules.

\subsection{Prompt directives}
\label{sec:promptdirec}

We enumerate our hypotheses about extending the prompt with prompt rules in the following list. 


\begin{enumerate}
    \item The performance of code generation can be improved by adding domain-specific rules $\rho_i$, where $\rho_i \in R$, to the prompt. These rules target a subset of questions, and generating the correct answer requires instructional constraints or external knowledge.
    \item It is possible to algorithmically define such a subset of errors that has a common root cause of failure 
    \item It is possible to introduce a  prompt rule that eliminates this root cause
    \item  It is possible to apply this rule globally, without additional decision logic within the agent. 
\end{enumerate}

 To prove these assumptions, we implemented Algorithm \ref{alg:pea}.
\begin{algorithm}
\caption{Prompt Extending Algorithm}
\label{alg:pea}
\begin{algorithmic}[1]

\Require Test dataset $D_{\text{test}} = \{(q_i,c_i,p_i, y_i)\}_{i=1}^n$
\State $R \gets \emptyset$

\While{\textbf{true}}
    \State $\hat{Y} \gets \textsc{Predict}(D_{\text{test}}, R)$
    \State $EM \gets \textsc{Score}(\hat{Y}, Y)$
    \State $E \gets \{ (q_i,c_i,p_i,y_i) \mid EM_i < 1 \}$
    \State $F \gets \textsc{ExtractFeatures}(E)$
    \State $(m_1, m_2) \gets \textsc{HyperparameterTuning}(F)$
    \State $\mathcal{K} \gets \textsc{HDBSCAN}(F, m_1, m_2)$
    \State $\mathcal{K}_{sorted} \gets \textsc{SortBySizeDesc}(\mathcal{K})$
    \State $i \gets 1$

    \While{$i \leq 2$}
        \State $\rho \gets \textsc{FormulateRule}(\kappa_{sorted,i})$
        \State $\hat{Y_C} \gets \textsc{Predict}(E, R \cup \{\rho\})$
        \State $EM_C \gets \textsc{Score}(\hat{Y_C}, Y_C)$
        \State $EM_{pairs} \gets \textsc{AlignPairs}(Y_C, Y, EM_C, EM)$
        \State $p_{exact} \gets \textsc{McNemarTestBinom}(EM_{pairs})$
        \State $\Delta EM \gets \sum_{j} EM_{C,j}$

        \If{$\Delta EM > 50\%$ \textbf{and} $p_{exact} <= 0.0625$}
            \State $R \gets R \cup \{\rho\}$
            \State \textbf{break} \Comment{re-run outer loop}
        \Else
            \State $i \gets i+1$
        \EndIf
    \EndWhile

    \If{$i > 2$}
        \State \textbf{break} \Comment{stop further search}
    \EndIf
\EndWhile

\State \Return $R$ \Comment{final rule set}

\end{algorithmic}
\end{algorithm}

Note pseudo function \textsc{FormulateRule}. Here a human-in-the-loop intervention is required, where the expert formulates a domain-specific prompt rule (DSPR) to address the identified error cluster.

\section{Implementation details}
\label{details}
We used the OLlama running environment, which was accessed via the OLlama API interface from Jupyter Notebook. Ollama version was 0.11.4. Agent source code and a demo application can be found in \cite{codgen_ollama}.

We selected model Qwen3 4B, the top performer in the previous stage. We used the version tagged with \texttt{qwen3:4b-q4\_K\_M}, id: \texttt{2bfd38a7daaf}. Its size is 2.6 GB, the context window size is 40k, and it uses a 4-bit, K-series, mixed quantization scheme. In this version, the thinking and non-thinking modes were not yet separated into two separate models, and the thinking behavior was controlled with the \texttt{/no\_think} flag within the prompt. 

\subsection{Feature Extraction}
For clustering, we used HDBSCAN \cite{McInnes2017} with a feature matrix. We extracted the following features:

\begin{itemize}
    \item \texttt{calc\_pattern}
    \item \texttt{code\_calc\_pattern}
    \item \texttt{scale\_mismatch}
    \item \texttt{value\_match}
    \item \texttt{error\_code}
\end{itemize}

\subsubsection{\texttt{calc\_pattern}}
The calculation pattern describes the type of arithmetic calculation needed to answer the question. We generate this value from the \texttt{derivation} property of the question, which contains the exact reference arithmetic calculation required to answer the question. To create groups from the same expressions, we replaced the numbers with a '\#' tag and performed additional data cleaning, such as removing extra parentheses and whitespace. We also group rare  (frequency < 3 ) values into the category of 'other' to prevent fragmentation. The frequency of these calculation patterns is shown in Table \ref{tab:calc_patterns}.

\begin{table}[ht!]
\caption{Frequency of calculation patterns.}
\label{tab:calc_patterns}
\centering
\small
\begin{tabular}{|l|r|}
\hline
Calculation pattern               & Frequency \\ \hline
\#-\#                             & 134       \\ \hline
(\#-\#)/\#                        & 100       \\ \hline
(\#+\#)/\#                        & 89        \\ \hline
\#/\#                             & 49        \\ \hline
(\#+\#+\#)/\#                     & 26        \\ \hline
\#+\#                             & 21        \\ \hline
\#/\#-\#                          & 9         \\ \hline
-\#-\#                            & 8         \\ \hline
\#\%-\#\%                         & 7         \\ \hline
-(\#+\#)/\#                       & 5         \\ \hline
(\#\%+\#\%)/\#                    & 5         \\ \hline
\#+\#+\#                          & 4         \\ \hline
{[}(\#+\#)/\#{]}-{[}(\#+\#)/\#{]} & 4         \\ \hline
other                             & 36        \\ \hline
\end{tabular}
\end{table}

\subsubsection{\texttt{code\_calc\_pattern}}
This feature has a similar format to \texttt{calc\_pattern}, but its extraction is different. Here, we analyze the abstract syntax tree of the generated code and extract the arithmetic calculations, including summarization in loops. We also use the "other" category here. 

\subsubsection{\texttt{scale\_mismatch}}
A simple Boolean value is true if the predicted and ground-truth scales differ in value. 

\subsubsection{\texttt{value\_match}}
This feature compares the predicted and ground-truth numeric value. 
\begin{itemize}
    \item 0: the two values are equal
    \item 1: sign error, the two values differ only in sign
    \item 2: predicted value is NaN  (usually indicates a runtime error)
    \item 3: else branch, the two values differ in value
\end{itemize}

\subsubsection{\texttt{error\_type}} 
The type of error categorizes the prediction errors based on the prediction results. We defined six types of error:

\begin{itemize}
    \item \textit{Selection Error}: the generated code does not find the expected number values in the table, according to the \textit{derivation} field in the test dataset.
    \item \textit{Calculation Error}: the generated code has found the expected number values in the table,  according to the \textit{derivation} field in the test dataset, but the calculation with these values failed.
    \item \textit{Scale Error}: the generated code managed to calculate the expected value of the number, but responded with an incorrect scale result.
    \item \textit{Sign Error}: the generated code managed to calculate the result of the expected number value, but with an incorrect sign.
    \item \textit{Syntax Error}: the generated code contains a syntax error.
    \item \textit{Runtime Error}: the generated code resulted in a runtime error.        
\end{itemize}

The feature matrix has been generated by a \texttt{DictVectorizer}, parametrized to create a dense matrix. As the features are binary or one-hot encoded categorical values, we selected the Hamming distance as a metric for distance calculation.


\[
d_H(x,y) = \left| \{\, i \in \{1,\dots,n\} \mid x_i \ne y_i \,\} \right|
\]
\[
d_H(x,y) = \sum_{i=1}^n \mathbf{1}[x_i \ne y_i]
\]

\subsection{Hyperparameter Tuning}

We implemented an exploration step to find an optimal value for the HDBSCAN parameters \texttt{min\_cluster\_size} and \texttt{min\_samples}. In this step, we calculate the cluster count and noise ratio for the permutations of \texttt{min\_cluster\_size = [2,3,4,5,6,10,15]} and \texttt{min\_samples = [1,2,3,4,5]}. Then we sort the values by noise ratio, and select the cluster count close to the first quartile (Q1) of the error set size. This count range was chosen following an investigation into the usability of the generated clusters. Here, we must consider that a small cluster count (large cluster sizes) combines several root causes into one cluster, whereas a large cluster count splits errors with a common root cause into separate clusters. Although we executed this exploration after each iteration in our experiments, the parameter pairs \texttt{min\_cluster\_size = 2} and \texttt{min\_samples = 1} were selected in each iteration. 
 
\subsection{Clustering}
After selecting clustering parameters, we run the HDBSCAN to cluster the errors. It yields a list of error clusters. We selected the most significant cluster $\kappa^\ast$ (having the maximal size) for investigation and to formulate a new prompt rule as described in the Prompt Extending Algorithm above. 

\subsection{Extended Prompt Performance Evaluation }

First, using the new prompt, we calculate the $EM$ values for the questions that were members of the selected error cluster. The improvement can be easily calculated by averaging the new $EM$ values,
\[
\Delta EM = \sum_{i=1}^n EM_{extended, i} / n-  \sum_{i=1}^n EM_{reference, i}  /n
\]
and we can simplify to 
\[
\Delta EM = \sum_{i=1}^n EM_{extended, i}
\]
as for each question in an error cluster $\kappa^\ast$, the Exact Match (EM) score 
is equal to zero:
\[
\forall (q, c, y) \in \kappa^\ast : \text{EM}(q, c, y) = 0,
\]
which follows directly from the fact that an error cluster contains only 
failed test instances. After that, we formed $EM$ value pairs from old and new predictions within the cluster, using the question ID as the key value. 

To compare these pairs and measure the significance of the improvement in $EM$ values, we used the McNemar test \cite{McNemar_1947} on these pairs. We chose the McNemar test because it is suitable for binary-value pairs and can provide a significance value $p$ to use as acceptance criteria for prompt variants.  We used the exact binomial variant of the McNemar test due to the low $b$ and $c$ values resulting from the small cluster size. We used $p <= 0.0625$ and $\Delta EM >= 0.5$ for acceptance criteria. We used a cross-tab heatmap of the calculation pattern by error type to visualize a prompt's error distribution.

\section{Experiments}
\label{sec:experiments}
\begin{figure}[ht!]
    \centering
    \includegraphics[width=1.0\linewidth]{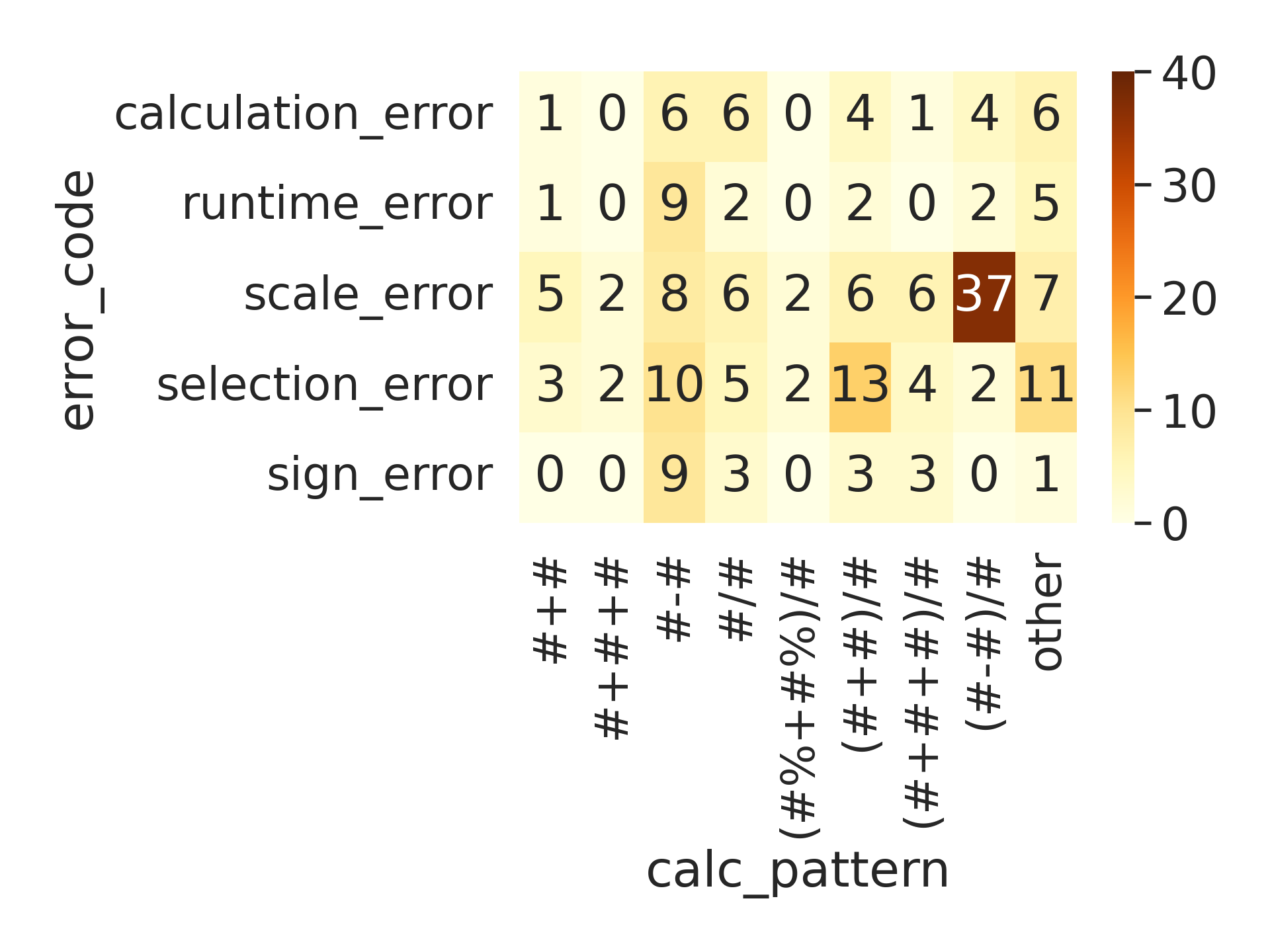}
    \caption{Cross-tab heatmap of the calculation pattern on error type, when no rules are present in the prompt.}
    \label{fig:heatmapv18}
\end{figure}

The following section describes the algorithm's usage on the TAT-QA dataset. We begin with a test run using the base prompt, which includes no specific rule guidelines. It can be seen in Appendix \ref{sec:baseprompt}. The heatmap for this prompt version is shown in Figure \ref{fig:heatmapv18}.

As a next step, we find an optimal set of HDBSCAN clustering parameters. We achieve it by running a grid search on \texttt{min\_cluster\_size }and \texttt{min\_samples} parameters, and sorting the result by noise ratio. Such an output can be seen on Table \ref{tab:grid_search}.

\begin{table}[h]
    \centering
    \footnotesize
    \begin{tabular}{|c|c|c|c|}
        \hline
         min\_cluster\_size& min\_samples&Cluster Count& Noise Ratio\\ \hline
         15 & 3 & 4  & 0.130653 \\ \hline
         2  & 1 & 45 & 0.140704 \\ \hline
         15 & 2 & 4  & 0.140704 \\ \hline
         10 & 2 & 5  & 0.140704 \\ \hline
         15 & 4 & 3  & 0.145729 \\ \hline
         3  & 1 & 25 & 0.165829 \\ \hline
         10 & 4 & 6  & 0.195980 \\ \hline
         10 & 3 & 6  & 0.195980 \\ \hline
         4  & 1 & 19 & 0.206030 \\ \hline
         15 & 1 & 4 & 0.216080  \\ \hline
    \end{tabular}
    \caption{The first 10 HDBSCAN clustering setups as a result of a grid search }
    \label{tab:grid_search}
\end{table}
As the error set contains 199 items, the cluster count using parameters (2,1) is the closest to 50, which is the Q1 quartile of the error count. So we proceed with this setup. Clusters can be seen in Table \ref{tab:cluster_item_counts}.

\begin{table}[h]
    \centering
    \small
    \begin{tabular}{|c|c|}
        \hline
        Cluster ID & Item Count \\ \hline
        24 & 34 \\ \hline
        -1 & 31 \\ \hline
         41& 8  \\ \hline
         6 & 7  \\ \hline
         9 & 5  \\ \hline
    \end{tabular}
    \caption{Item counts per cluster for the first 5 clusters}
    \label{tab:cluster_item_counts}
\end{table}

\begin{figure}[ht!]
    \centering
    \includegraphics[width=1.0\linewidth]{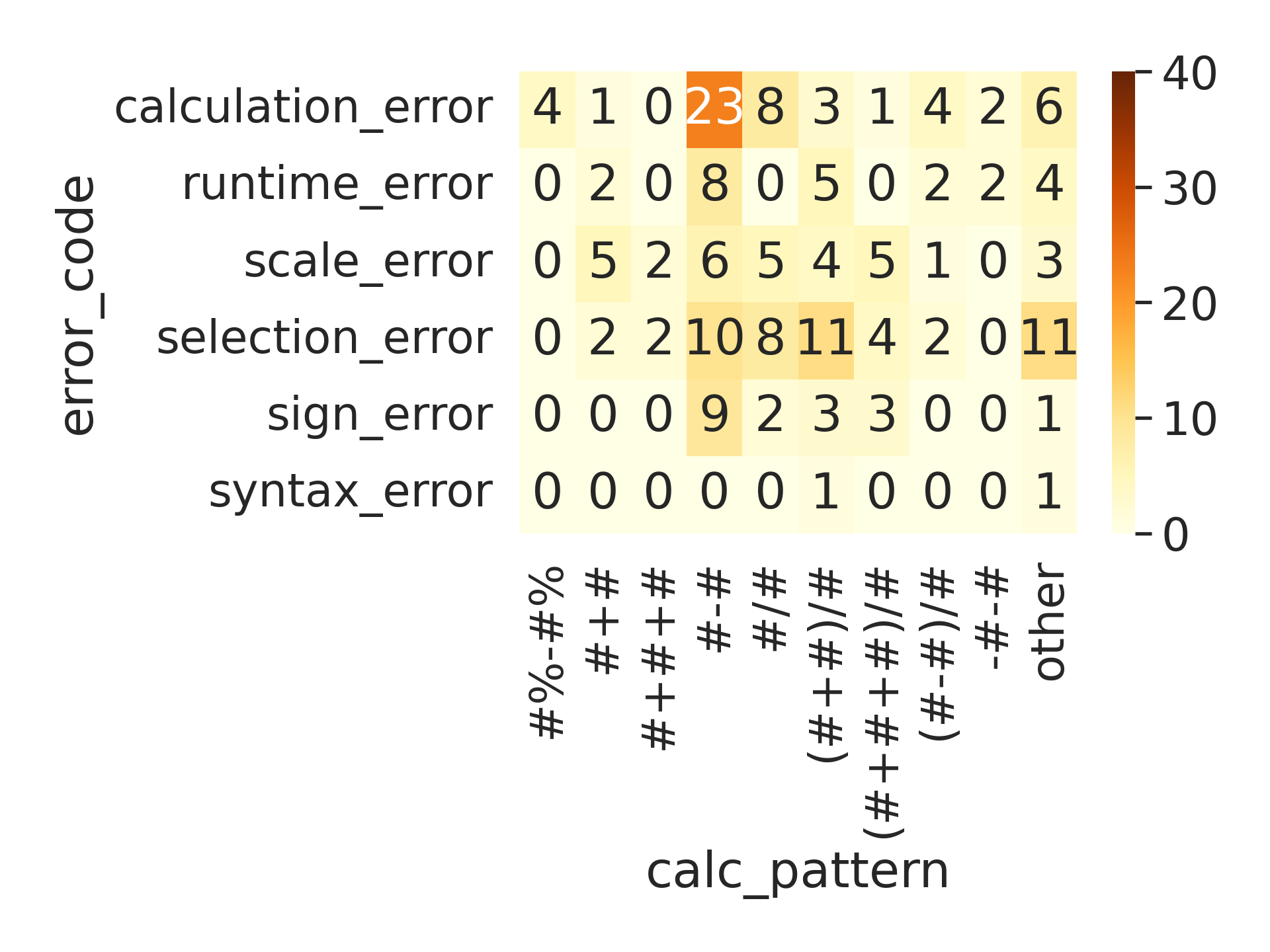}
   \caption{Cross-tab heatmap of the calculation pattern on error type, when one rule is added into the prompt that forces a percent scale for percentage change calculations.}
    \label{fig:heatmapv23}
\end{figure}

We select the first cluster (ID:24), with the largest item count of 34. The error subset of this cluster can be seen in Appendix \ref{sec:errcluster24}. We can see that all the errors are scale errors, where the scale should be 'percent', as the question is about 'percentage change'.  So we extend the base prompt with the new rule \textit{"'percentage change' results 'percent' scale"}. We can run tests on the error cluster using the new prompt to measure its local performance. The binomial McNemar test yielded a p-value of $2.15E-05$ and a $\Delta EM$ of $90.9\%$. Thus, we accept the prompt and calculate the global $EM$, which is 64.59\%. Global $\Delta EM = 4.63\%$. According to this result, we can also accept the new prompt globally. The crass-tab heatmap of the new run can be seen in Figure \ref{fig:heatmapv23}. 

It can be seen that the center of gravity of errors moved from the percentage change calculation to difference one, while the global error decreased.

\begin{figure}[ht!]
    \centering
    \includegraphics[width=1.0\linewidth]{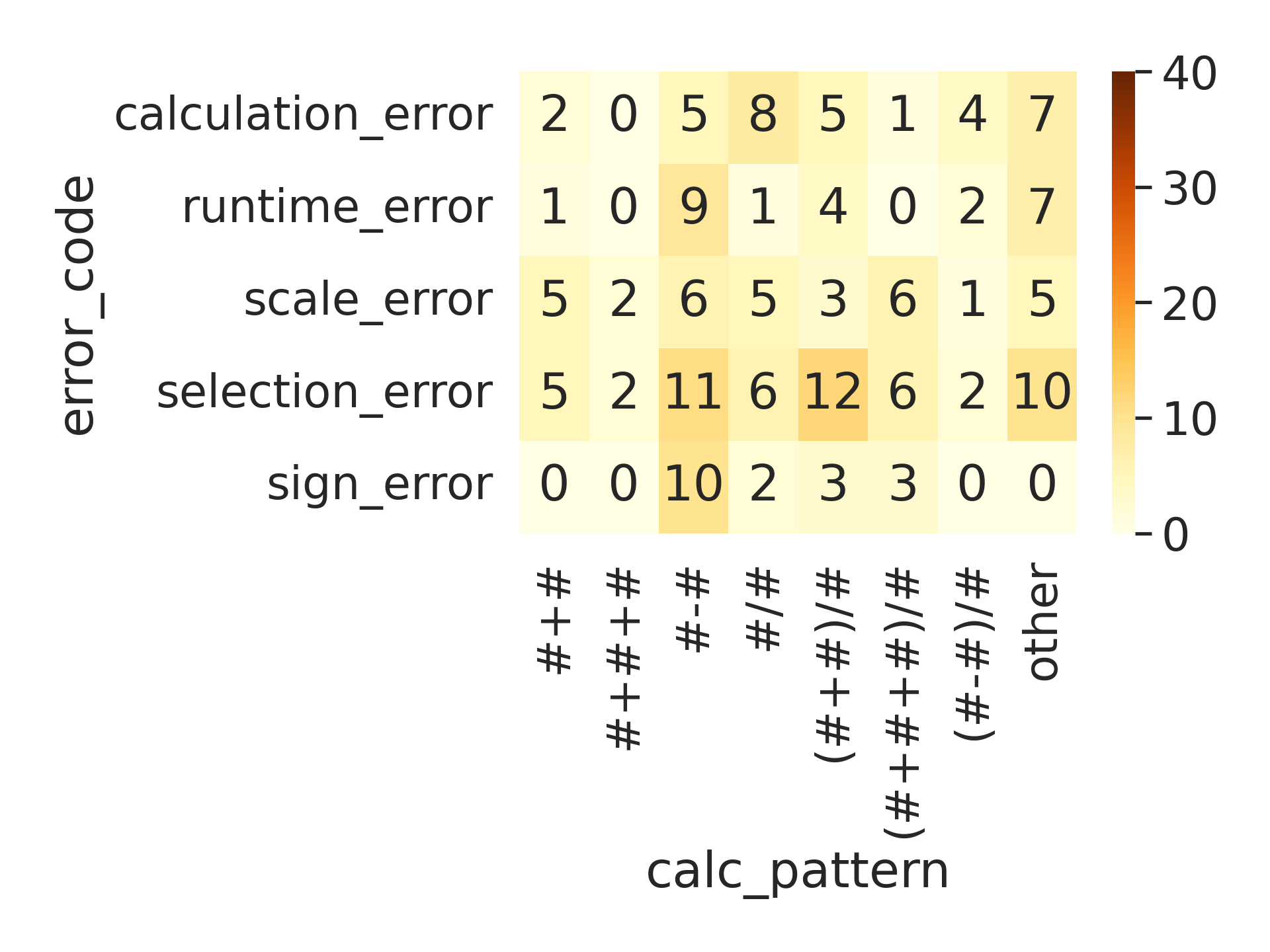}   
   \caption{Cross-tab heatmap of the calculation pattern on error type, when a rule is added to calculate the change in percentage as subtraction.}
    \label{fig:heatmapv41}
\end{figure}

In the second iteration, we create the error clusters the same way as above. Here, we select the parameter pair \texttt{min\_cluster\_size = 2} and \texttt{min\_samples = 1} again, resulting in a cluster count of 43 and a noise ratio of 0.164773. After clustering and sorting by cluster size, we select the largest cluster, which has 22 items and a size of 8. We can note that the cluster sizes start to drop after the first large cluster, which grouped scale mismatch as the root cause. The selected cluster shows a root case in which the language model handles 'change in percentage' type questions as a relative change instead of a subtraction. Using this knowledge, we can formulate a new rule: \textit{"'change in percentage' is a subtraction"}. The McNemar test yielded a p-value of $0.0265$ and a $\Delta EM$ of $62.50\%$. We accept the prompt and calculate the global $EM$, which is 67.61\%. Global $\Delta EM = 3.02\%$. According to this result, we can also accept the new prompt on a global level. The crass-tab heatmap of the new run can be seen in Figure \ref{fig:heatmapv41}. 

After selecting the largest error cluster, we identified the root cause: the order of the minuend and subtrahend when calculating the difference. New rule \textit{"If the question is about the 'difference' or 'change' between two values, always subtract the first from the second"} yields p-value of $0.25$ and a $\Delta EM$ of $30.50\%$. We drop this rule because it is below the algorithm's limit.

\begin{figure}[ht!]
    \centering
    \includegraphics[width=1.0\linewidth]{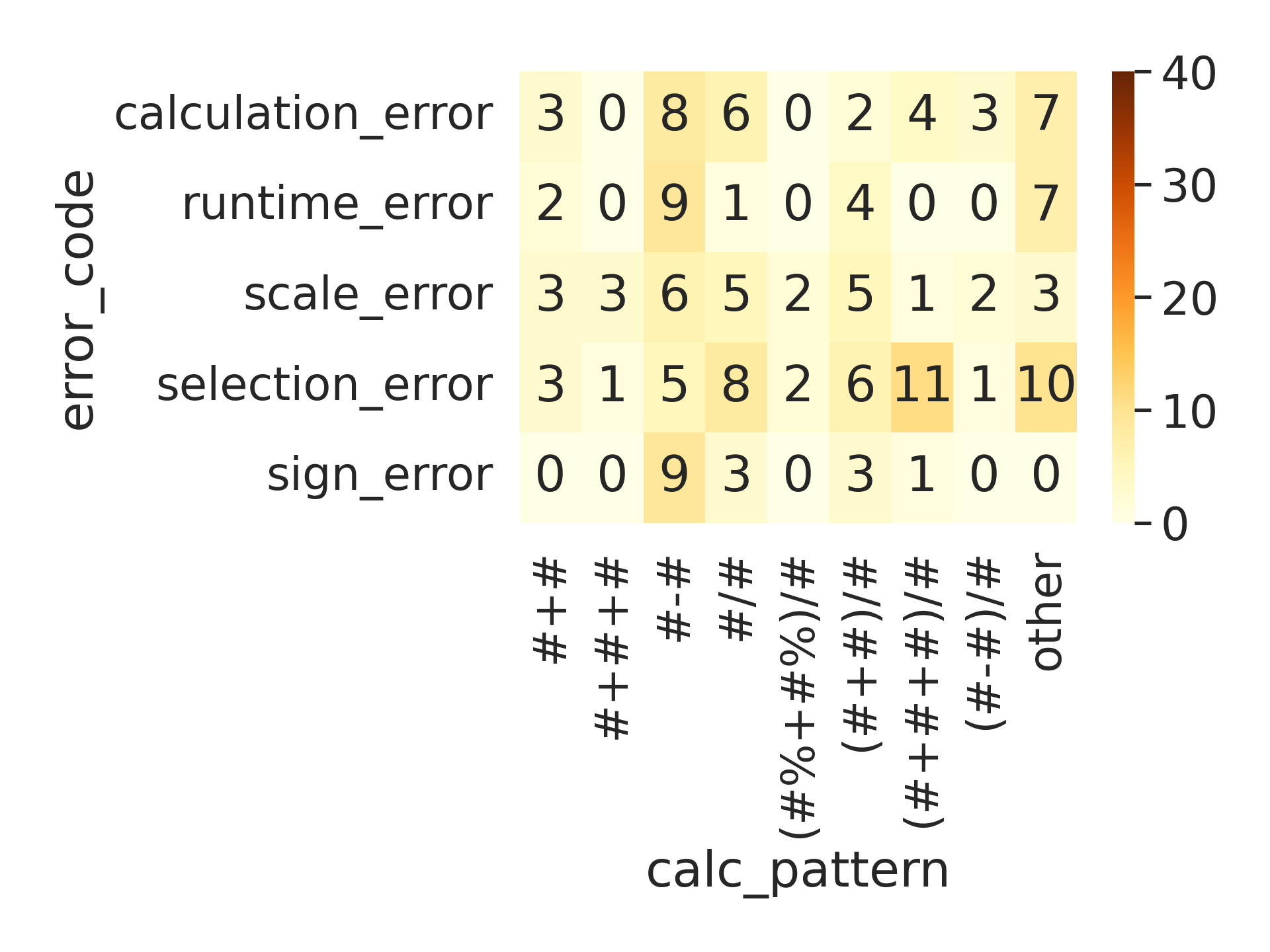}
 \caption{Cross-tab heatmap of the calculation pattern on error type, when a rule is added to calculate the year average into the prompt.}
    \label{fig:heatmapv29}
\end{figure}

After that, we select the cluster with ID 22 from the previous cluster set. We identify the root cause: the financial term "year average" is unknown to the language model. Thus, we introduce a new rule \textit{"If the question is about calculating the year average, you
must calculate the average between the given year and
the previous one. Ex. 2015 average = (2015 value +
2014 value)/2."}, which yields a p-value of $0.0625$ and a $\Delta EM$ of $83.33\%$. Now we can accept the prompt, with global $EM = 70.82\%$ and global $\Delta EM = 3.22\%$. The crass-tab heatmap of the new run can be seen in Figure \ref{fig:heatmapv29}. 

After new clustering and selecting the largest error cluster, the cluster contains runtime errors where the category or header selection had typos or generated code used incorrect JSON identifiers during the selection process. We can formulate a new rule: \textit{"Use exact category or header value to select a number value; number value is in property 'number\_value'"}, which yielded p-value of $0.5$ and a $\Delta EM$ of $28.57\%$. We drop this rule because it is below the algorithm's limit. 

In the next iteration, after selecting the next largest error cluster, the cluster contains errors where the average calculation of more than two values fails, as the generated code uses only two values for averages. We can formulate a new rule: \textit{"To calculate the average, use all relevant years"}, which yields a p-value of $0.125$ and a $\Delta EM$ of $57.14\%$. We drop this rule because it is below the algorithm's limit. (Considering the relatively high $\Delta EM$, we can calculate the global $\Delta EM$, which is -0. 8\% in this case).

As we reach our stop criteria, the algorithm is stopped, and the return value is the base prompt of the last iteration.  The final prompt contains the following rules:
\begin{itemize}
    \item 'percentage change' results 'percent' scale
    \item 'change in percentage' is a subtraction
    \item If the question is about calculating the year average, you must calculate the average between the given year and the previous one.  Ex. 2015\_average = (2015\_value + 2014\_value)/2.
\end{itemize}
The full final prompt can be seen in Appendix \ref{sec:finalprompt}

\section{Results}
\label{sec:results}

\subsection{Measurement results}

\begin{figure}[ht!]
    \centering
    \includegraphics[width=1.0\linewidth]{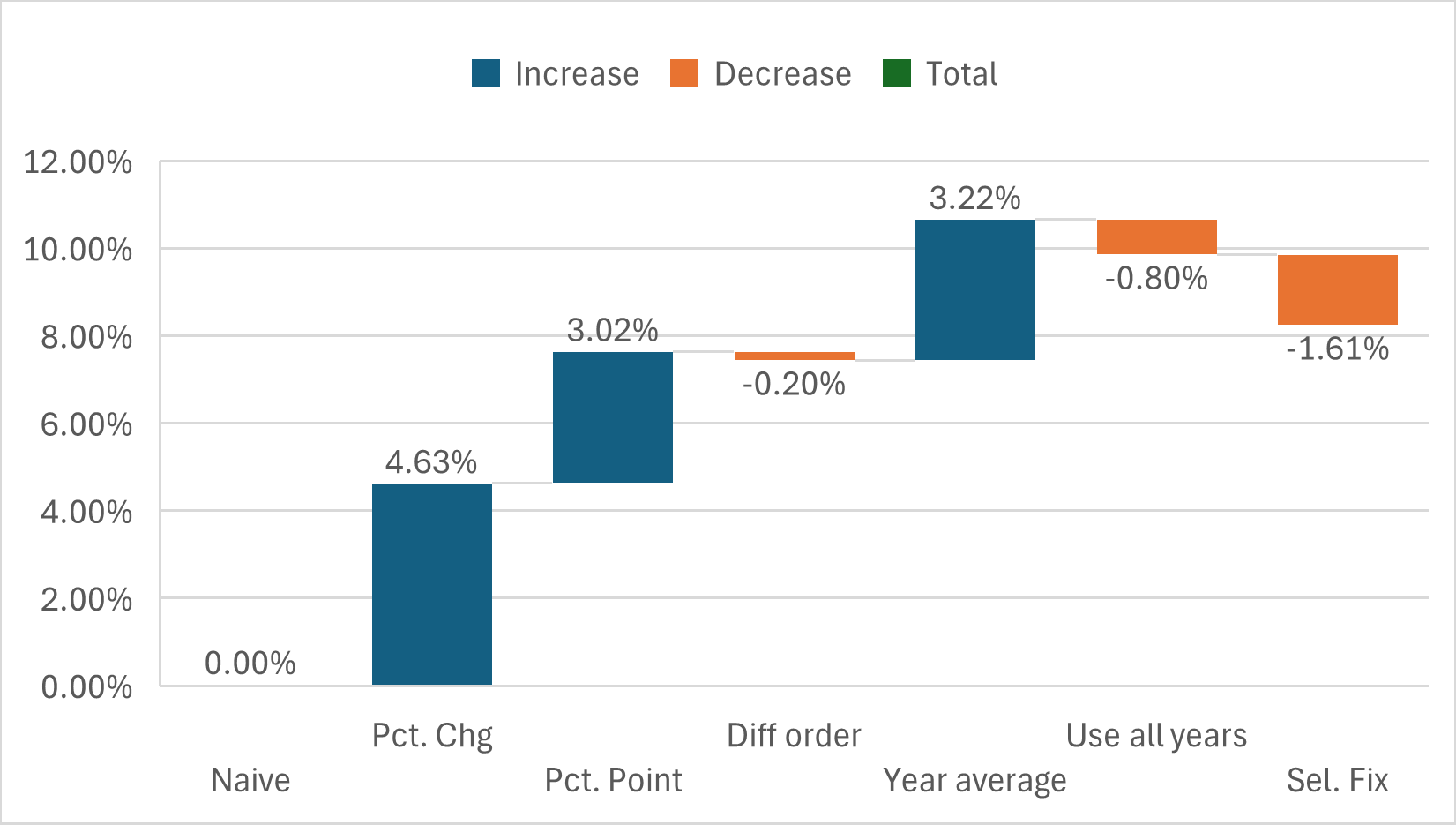}
 \caption{Prompt Extending Algorithm iterations with rules.
Impact of rules on global performance}
    \label{fig:promptchart}
\end{figure}

We investigated the effect of transferring the original prompt rules, used in large language models, to the SLM prompt to further improve the model performance \ref{sec:experiments}. Compared to our paper \cite{pandy_ines_2025}, the main result is that we defined and presented the performance of a well-defined algorithm to introduce new prompt rules, which also has the advantage that it can be automated later.  For this experiment, we selected the best model, Qwen3 4B, from the previous work. The results are presented in Table \ref{tab:prompts} and Table \ref{tab:rules2}. In the first row of the table, we can see the performance of the simplified prompt without added rules. The results of the experiments described in Section \ref{sec:experiments} are presented in the following rows. Figure \ref{fig:promptchart} shows the graphical representation of reaching the optimal maximum rule count during iterations. 

\begin{table}[]
\caption{Prompt Extending Algorithm iterations summary.  Accepted versions are in bold.}
\label{tab:prompts}
\footnotesize
\begin{tabular}{|l|l|r|r|r|r|r|r|}
\hline
 &  & \multicolumn{4}{c|}{McNemar Test} &  &  \\ \hline
\multicolumn{1}{|c}{Prompt} & \multicolumn{1}{|c}{Base}   & \multicolumn{1}{|c}{N} & \multicolumn{1}{|c}{+} & \multicolumn{1}{|c}{$\Delta$} & \multicolumn{1}{|c}{p} & \multicolumn{1}{|c}{EM} & \multicolumn{1}{|c|}{$\Delta$} \\ \hline
\textbf{V1} &  &  & - &  &  & \textbf{59.96\%} &  \\ \hline
\textbf{V2} & \textbf{V1} & \textbf{22} & \textbf{20} & \textbf{90.91\%} & \textbf{2.1E-05} & \textbf{64.59\%} & \textbf{4.63\%} \\ \hline
\textbf{V3} & \textbf{V2} & \textbf{8} & \textbf{5} & \textbf{62.50\%} & \textbf{0.0625} & \textbf{67.61\%} & \textbf{3.02\%} \\ \hline
V4 & V2 & 10 & 3 & 30.00\% & 0.2500 & 67.40\% & -0.20\% \\ \hline
\textbf{V5} & \textbf{V4} & \textbf{6} & \textbf{5} & \textbf{83.33\%} & \textbf{0.0625} & \textbf{70.82\%} & \textbf{3.22\%} \\ \hline
V6 & V5 & 7 & 4 & 57.14\% & 0.1250 & 70.02\% & -0.80\% \\ \hline
V7 & V5 & 7 & 2 & 28.57\% & 0.5000 & 69.22\% & -1.61\% \\ \hline
\end{tabular}
\end{table}

\begin{table}[]
\caption{Prompt Extending Algorithm iterations with rules. Accepted versions are in bold.}
\label{tab:rules2}
\footnotesize
\begin{tabular}{|l|l|l|}
\hline
ID & Prompt Rules & EM\%\\ \hline
V1 & - & 59.96\\ \hline
V2 & - 'percentage change' results 'percent'   scale & 64.59\\ \hline
V3 & \begin{tabular}[c]{@{}l@{}} - percentage   change' results 'percent' scale; \\ - 'change in percentage' is a subtraction;\end{tabular}  & 67.61\\ \hline
V4 & \begin{tabular}[c]{@{}l@{}}- percentage change' results 'percent' scale;\\ - 'change in percentage' is a subtraction;          \\ - If the question is about the 'difference' or 'change' \\ between two values, \\ always subtract the first from the second;\end{tabular} & 67.40\\ \hline
V5 & \begin{tabular}[c]{@{}l@{}}- If the question is about calculating \\the year average, you must calculate \\ the average between the given year \\ and the previous one.\\    ex. 2015\_average =   (2015\_value + 2014\_value)/2\\ - 'percentage change' results 'percent' scale\\ - 'change in percentage' is a   subtraction\end{tabular} & 70.82\\ \hline
V6 & \begin{tabular}[c]{@{}l@{}}'percentage change' results 'percent' scale\\ 'change in percentage' is a   subtraction\\ To calculate the average, use all relevant years.;\end{tabular} & 70.02\\ \hline
V7 & \begin{tabular}[c]{@{}l@{}}- 'percentage change' results 'percent' scale;         \\ - 'change in percentage' is a subtraction;       \\ - Use exact category or header value to \\ select a number value; the number value is\\ in the property 'number\_value';\end{tabular} & 69.22\\ \hline
\end{tabular}
\end{table}

\begin{table}[ht!]
\centering
\caption{Comparison of a large and a small language model's performance, using CGA and prompt rules}
\label{tab:largesmall}
\small
\begin{tabular}{|l|r|r|}
\hline
\multicolumn{1}{|c|}{\multirow{2}{*}{Model}} & \multicolumn{1}{c|}{\multirow{2}{*}{Exact Match}} & \multicolumn{1}{c|}{\multirow{2}{*}{Value Match}} \\
\multicolumn{1}{|c|}{} & \multicolumn{1}{c|}{} & \multicolumn{1}{c|}{} \\ \hline
\multirow{2}{*}{GPT 3.5 turbo} & \multirow{2}{*}{66.27\%} & \multirow{2}{*}{\textbf{78.92\%}} \\
 &  &  \\ \hline
\multirow{2}{*}{Qwen3 4B} & \multirow{2}{*}{\textbf{70.82\%}} & \multirow{2}{*}{77.06\%} \\
 &  &  \\ \hline
\multirow{2}{*}{Change} & \multirow{2}{*}{4.55\%} & \multirow{2}{*}{-1.86\%} \\
 &  &  \\ \hline
\end{tabular}
\end{table}

Table \ref{tab:largesmall} compares the final model Qwen3 4B setup and our reference model setup with GPT-3.5 Turbo. Both setups apply the same features: table restructuring, Code Generation Agent, and model-specific prompt rules. It shows that small models' performance is comparable to larger models if we apply the Code Generation Agent and appropriate prompt rules.

An interesting finding in this comparison is that while the Qwen3 4B agent achieves parity in overall Exact Match, its performance on the numerical value alone (w/o scale accuracy) remains slightly lower than that of GPT-3.5 Turbo (78.92\% vs 77.06\%). This suggests that model-specific rule tuning can effectively guide the SLM in producing the correct final format (including the scale). However, the underlying core reasoning and arithmetic selection may still be less robust than in a larger model. This highlights a potential trade-off between the efficiency of SLMs and their reliance on complex, multi-step numerical reasoning.

\subsection{Theoretical implication}

Let $K$ denote the number of prompt rules included in the prompt.
Small language models (SLMs) have limited capacity to process complex
instructions, and empirical evidence suggests they perform better with more straightforward,
more concise prompts. When $K$ becomes large, the model may be "overloaded",
increasing the risk of misinterpreting or ignoring some rules.

Furthermore, as $K$ increases, the probability of \emph{rule conflict} also
increases, introducing ambiguity into the task. The marginal gain in accuracy
from each additional rule can be modeled as $\varepsilon(K)$, where:
\[
\varepsilon(K) \quad\text{satisfies}\quad \varepsilon(K) > 0 \ \text{for} \ K < K_{\text{opt}},
\]
and 
\[
\varepsilon(K) \leq 0 \ \text{for} \ K \geq K_{\text{opt}}.
\]
The first few rules fix common errors (large $\varepsilon$), while later rules
only address rare edge cases (small $\varepsilon$), and may even harm accuracy.

Thus, there exists an optimal number of rules $K_{\text{opt}}$ that maximizes
performance:
\[
K_{\text{opt}} = \arg\max_{K} \ P_{\text{accuracy}}(K).
\]
Beyond $K_{\text{opt}}$, performance stagnates or decreases, implying that rule
selection should be guided by data-driven fine-tuning rather than by simply
maximizing $K$.

\subsection{Application}

Using our presented results, we have created a specialized and containerized agent for processing tabular data based on the Qwen model \cite{qwen3technicalreport}. Due to containerization, our solution is simple to deploy, robust, and easily scalable.

\section{Conclusion}
\label{sec:conclusion}





The perceived arithmetic limitations of small, locally deployable language models (SLMs) are not fundamental, but rather an artifact of the inference strategy. This research demonstrates that decomposing complex, multi-step tabular reasoning into deterministic, verifiable code generation fundamentally alters the performance landscape.

However, task decomposition alone is insufficient. The critical breakthrough comes from introducing a systematic, error-driven optimization cycle: algorithmic clustering of the model's prediction errors allows for the precise identification of error root causes (such as the misunderstanding of "percentage change" or "year average") and their targeted remediation through domain-specific prompt rules.

The fusion of these two methods—code decomposition and iterative, error-driven refinement—is potent enough to elevate the accuracy of a compact 4B-parameter model (Qwen3 4B) from 59.96\% to 70.82\%, thereby surpassing the performance of the significantly larger, general-purpose GPT-3.5 Turbo (66.27\%). This finding breaks the dependency on large, API-based models for highly sensitive, privacy-constrained domains such as financial analysis.

Our results also outline a new optimization paradigm: peak performance is achieved not by the accumulation of rules, but by finding an optimal, concise rule set ($K_{\text{opt}}$), beyond which cognitive overload degrades the results. This confirms that SLM optimization is a distinct discipline, not a simple downscaling of LLM strategies. The presented procedure serves as a blueprint for developing auditable, resource-efficient, and privacy-preserving analytical agents that operate entirely on-premises.

A critical open question remains, however, as to how these specialized small models, equipped with a fine-tuned rule-set, can generalize to new, unseen domains without the iterative error-analysis cycle presented here.

\clearpage
\section{Prompt Versions}

\subsection{Base Prompt}
\label{sec:baseprompt}
\textbf{system}: You will receive the financial report as an annotated value list and the question. 
Your task is to generate a Python function that can calculate a numeric value that is the answer for the received question. 

\textbf{human}: 

VALUE\_LIST: \{value\_list\}

QUESTION: \{question\}

Generate a Python function 'run(value\_list)' that can answer the question using the list of annotated values! 
The function must return a tuple (number, scale). The resulting number is a float with accuracy to two decimal places. Scale usually is thousand, million, billion, percent or an empty string. 

Do not generate explanation, nor example code, just the function. 

\subsection{Prompt with Rules - Final Prompt}
\label{sec:finalprompt}
\textbf{system}: You will receive the financial report as an annotated value list and the question. 
Your task is to generate a Python function that can calculate a numeric value that is the answer for the received question. 

\textbf{human}: 

VALUE\_LIST: \{value\_list\}

QUESTION: \{question\}

Generate a Python function 'run(value\_list)' that can answer the question using the list of annotated values! 
The function must return a tuple (number, scale). The resulting number is a float with accuracy to two decimal places. Scale usually is thousand, million, billion, percent or an empty string. 

\textit{If the question is about calculating the year average, you must calculate the average between the given year and the previous one. ex. 2015\_average = (2015\_value + 2014\_value)/2}

\textit{'percentage change' results 'percent' scale}

\textit{'change in percentage' is a subtraction}

Do not generate an explanation, nor example code, just the function. 

\clearpage
\onecolumn
\section{Error Cluster Members}
\label{sec:errcluster24}
\begin{table}[!htbp]
\label{tab:cluster24}
\footnotesize
\begin{tabular}{|l|p{7.5cm}|l|l|l|l|l|l|}
\hline
ID & question & calc pat. & code calc pat. & scale & pred scale & err. code & clus. \\ \hline
1 & What was the percentage change in the amount   for Appliances in 2019 from 2018? & (\#-\#)/\# & ((\#-\#)/\#)*\# & percent & million & scale\_error & 24 \\ \hline
31 & What was the percentage change in Total Other operating expenses between   2018 and 2019? & \textbf{(\#-\#)/\#} & ((\#-\#)/\#)*\# & percent & million & scale\_error & 24 \\ \hline
54 & What was the percentage change in Total Other in 2018 from 2017? & (\#-\#)/\# & ((\#-\#)/\#)*\# & percent &  & scale\_error & 24 \\ \hline
\textbf{67} & What was the percentage change in Adjusted EBITDA between 2018 and 2019? & \textbf{(\#-\#)/\#} & \textbf{((\#-\#)/\#)*\#} & percent &  & scale\_error & 24 \\ \hline
68 & What was the percentage change in the Total net deferred tax assets   between 2018 and 2019? & (\#-\#)/\# & ((\#-\#)/\#)*\# & percent &  & scale\_error & 24 \\ \hline
71 & What is the percentage increase in the number of rights 'outstanding at   the start of period' from 2018 to 2019? & (\#-\#)/\# & ((\#-\#)/\#)*\# & percent &  & scale\_error & 24 \\ \hline
86 & What is the percentage change in the value of raw materials between 2018   and 2019? & (\#-\#)/\# & ((\#-\#)/\#)*\# & percent & thousand & scale\_error & 24 \\ \hline
87 & What is the percentage change in the value of work-in-process inventory   between 2018 and 2019? & (\#-\#)/\# & ((\#-\#)/\#)*\# & percent & thousand & scale\_error & 24 \\ \hline
88 & What is the percentage change in the value of finished goods between 2018   and 2019? & (\#-\#)/\# & ((\#-\#)/\#)*\# & percent & thousand & scale\_error & 24 \\ \hline
113 & What was the percentage change in cash and cash equivalents from 2018 to   2019 year end? & (\#-\#)/\# & ((\#-\#)/\#)*\# & percent & thousand & scale\_error & 24 \\ \hline
182 & What is the percentage change in revenues from 2018 to 2019? & (\#-\#)/\# & ((\#-\#)/\#)*\# & percent & thousand & scale\_error & 24 \\ \hline
227 & What is the percentage change in the amounts owed to members of Peel from   2018 to 2019? & (\#-\#)/\# & ((\#-\#)/\#)*\# & percent & percent & scale\_error & 24 \\ \hline
252 & What was the percentage change in the average life expectancy for a male   member aged 65 in 2019 from 2018? & (\#-\#)/\# & ((\#-\#)/\#)*\# & percent &  & scale\_error & 24 \\ \hline
256 & What was the percentage change in the Total amount in 2019 from 2018? & (\#-\#)/\# & ((\#-\#)/\#)*\# & percent & million & scale\_error & 24 \\ \hline
273 & What is the percentage change in the total fair value of consideration   transferred at June 30 and December 31, 2019? & (\#-\#)/\# & ((\#-\#)/\#)*\# & percent &  & scale\_error & 24 \\ \hline
283 & What was the percentage change in Net income attributable to American   Tower Corporation stockholders between 2018 and 2019? & (\#-\#)/\# & ((\#-\#)/\#)*\# & percent & million & scale\_error & 24 \\ \hline
284 & What is the percentage increase in cash provided from financing   activities between 2018 and 2019? & (\#-\#)/\# & ((\#-\#)/\#)*\# & percent &  & scale\_error & 24 \\ \hline
285 & What is the percentage change in the total other non-current assets   between 2018 and 2019? & (\#-\#)/\# & ((\#-\#)/\#)*\# & percent &  & scale\_error & 24 \\ \hline
294 & What is the percentage change in the research and development costs   incurred from 2018 to 2019? & (\#-\#)/\# & ((\#-\#)/\#)*\# & percent &  & scale\_error & 24 \\ \hline
296 & What is the percentage change in the finance costs expensed from 2018 to   2019? & (\#-\#)/\# & ((\#-\#)/\#)*\# & percent &  & scale\_error & 24 \\ \hline
298 & What is the percentage change in cash provided by financing activities   between 2018 and 2019? & (\#-\#)/\# & ((\#-\#)/\#)*\# & percent &  & scale\_error & 24 \\ \hline
303 & What is the percentage change of net operating losses and credits from   2018 to 2019? & (\#-\#)/\# & ((\#-\#)/\#)*\# & percent &  & scale\_error & 24 \\ \hline
315 & What was the percentage change in Transportation Solutions in 2019 from   2018? & (\#-\#)/\# & ((\#-\#)/\#)*\# & percent & million & scale\_error & 24 \\ \hline
320 & What is the percentage change in the interim dividend from 2018 to 2019? & (\#-\#)/\# & ((\#-\#)/\#)*\# & percent & thousand & scale\_error & 24 \\ \hline
324 & What is the percentage change in Other for deferred tax assets? & (\#-\#)/\# & ((\#-\#)/\#)*\# & percent & million & scale\_error & 24 \\ \hline
336 & What was the percentage change in the amount Outstanding at 1 April in   2019 from 2018? & (\#-\#)/\# & ((\#-\#)/\#)*\# & percent &  & scale\_error & 24 \\ \hline
343 & What was the percentage change in the number of shares granted in 2019   from 2018? & (\#-\#)/\# & ((\#-\#)/\#)*\# & percent &  & scale\_error & 24 \\ \hline
367 & What was the percentage change in the amount Outstanding at 31 March in   2019 from 2018? & (\#-\#)/\# & ((\#-\#)/\#)*\# & percent &  & scale\_error & 24 \\ \hline
402 & What is the percentage change in net income between fiscal years 2019 and   2018? & (\#-\#)/\# & ((\#-\#)/\#)*\# & percent &  & scale\_error & 24 \\ \hline
409 & What is the percentage change in net cash provided by operating   activities between 2017 and 2018? & (\#-\#)/\# & ((\#-\#)/\#)*\# & percent &  & scale\_error & 24 \\ \hline
410 & What is the percentage change in the net cost of land, property, and   equipment in 2019 compared to 2018? & (\#-\#)/\# & ((\#-\#)/\#)*\# & percent & thousand & scale\_error & 24 \\ \hline
434 & What is the percentage change in gross deferred tax assets in 2019   compared to 2018? & (\#-\#)/\# & ((\#-\#)/\#)*\# & percent & thousand & scale\_error & 24 \\ \hline
475 & What is the percentage change of revenue from Hong Kong from 2017 to   2018, based on the geographic location of the customer's headquarters? & (\#-\#)/\# & ((\#-\#)/\#)*\# & percent &  & scale\_error & 24 \\ \hline
480 & What is the percentage change in the total balance of goodwill of the   Network Software \& Systems segment in 2019 compared to 2018? & (\#-\#)/\# & ((\#-\#)/\#)*\# & percent &  & scale\_error & 24 \\ \hline
\end{tabular}
\end{table}

\twocolumn
\newpage

\bibliographystyle{IEEEtranUrldate.bst}
\bibliography{ts}

\section*{Acknowledgment}
This work was supported by the project TKP2021-NKTA-34, which was implemented with support provided by the National Research, Development, and Innovation Fund of Hungary under the TKP2021-NKTA funding scheme. 

Furthermore, the authors utilized Grammarly and GPT-5 to enhance the clarity and coherence of the manuscript's language.

\end{document}